%% file: main.tex
\documentclass[a4paper]{article}
\pdfoutput=1
\usepackage{microtype}
\usepackage{graphicx}
\usepackage{subfigure}
\usepackage{booktabs} % for professional tables
\usepackage{amsmath}
\usepackage{multirow}
\usepackage{float}
\usepackage{amssymb,amsmath,epsfig}
\usepackage{scalerel}
\usepackage{microtype}
\usepackage{graphicx}
\usepackage{multirow}
\usepackage{booktabs}
\usepackage{amssymb,amsmath,graphicx}
\usepackage{caption}
\usepackage[export]{adjustbox}
\usepackage{moredefs}
\usepackage{color}\restylefloat{table}
\usepackage{enumitem}
\usepackage{hyperref}
\usepackage{multicol}

\usepackage{INTERSPEECH2020}

\title{Analyzing the Quality and Stability of a Streaming End-to-End On-Device Speech Recognizer}
\name{Yuan Shangguan*†\thanks{*Equal Contribution}\thanks{†This work was done while author was working at Google}, Kate Knister*, Yanzhang He, Ian McGraw, Françoise Beaufays}
\address{Google LLC, 1600 Amphitheatre Parkway, Mountain View, CA}
\email{yuansg@fb.com,kateknister@google.com}

\begin{document}
\maketitle

\begin{abstract}
\input{Abstract}
\end{abstract}

\section{Introduction}\label{Introduction}
\input{Intro}

\section{Instability Metrics}\label{Define}
\input{Define}

\section{Stability of the Streaming RNN-T}\label{RNNT}
\input{RNNT}

\section{Improving Text Normalization Stability}\label{TextNorm}
\input{TextNorm}

\section{Improving Streaming Stability}\label{Stream}
\input{Stream}

\section{Conclusion}\label{conclude}
\input{Conclude}

\section{Acknowledgements}
We thank Arun Narayanan and Tara Sainath for their efforts in
developing the RNN-T models with multi-domain training data,
text normalization of training data, and the domain-id inputs. We also appreciate
Dirk Padfield's careful review of our drafts.

\bibliographystyle{IEEEtran}
\bibliography{main}

\end{document}

%% file: Abstract.tex
The demand for fast and accurate incremental speech
recognition increases as the applications of automatic speech recognition (ASR)
proliferate. Incremental speech recognizers output chunks of partially recognized
words while the user is still talking. Partial results can be revised before
the ASR finalizes its hypothesis, causing instability issues. We analyze the
quality and stability of on-device streaming end-to-end (E2E) ASR models. We
first introduce a novel set of metrics that quantify the instability at word
and segment levels. We study the impact of several model training techniques
that improve E2E model qualities but degrade model stability.
We categorize the causes of instability and explore various solutions
to mitigate them in a streaming E2E ASR system.

%% file: Intro.tex
Modern applications of automatic speech recognition have brought to
users fast and accurate incremental speech recognition experiences.
These speech recognizers stream chunks of partially recognized
words to the user interface while the user is still talking.
We hereby refer to these chunks as~\textit{partials}.
Not only do users see text appear in real-time before the ASR recognizer finalizes
the transcription, downstream models such as
spoken dialog systems~\cite{selfridge2011stability}, real-time translation
system~\cite{arivazhagan2019re} and multimodal user
interfaces~\cite{fink1998incremental} also rely on partials to reduce
overall application latencies.

Prior works in streaming ASR highlight a significant obstacle:
model stability~\cite{selfridge2011stability,baumann2009assessing,mcgraw2012estimating}.
Recognizers often change emitted words before finalizing the hypothesis. These revisions
of partials cause flicker on the speech user interface. They also increase the
cognitive load of the users, distracting them from speaking and resulting in frustration.
Revisions also cause the downstream models to repeat operations, increasing overall application latency.

In this paper, we analyze the stability issues of an on-device,
streaming, Recurrent Neural Network Transducer (RNN-T) ASR recognizer.
It is a state-of-the-art E2E ASR recognizer that produces
transcripts with low word error rates (WER) on devices while staying
within the constraints of latency, memory storage and computational
resources~\cite{yuan2020}. We improve the training data diversity by
mixing data from multiple speech domains~\cite{narayanan2019recognizing,Sainath2020stream}.
Multi-domain training data improves robustness of the model for conditions not
seen during training, such as long-form audio
recognition~\cite{narayanan2018toward}. In this paper, we analyze the impact of
the diverse training data on the quality and stability of RNN-T recognizers.
Moreover, to avoid the additional latency from text normalizers,
we include capitalizations, spoken punctuations,
and Arabic numerals in the training data, so that the trained models output these written-text
formats~\cite{he2019streaming}.
Due to the diversity of transcript text formats present in the multi-domain training data,
the on-device RNN-T is especially prone to instability, revising its transcripts
frequently during the generation of partials.

Prior work proposed different methods for improving and evaluating stability of
traditional streaming ASR recognizers. Both~\cite{mcgraw2012estimating}
and~\cite{chen2016stable} defined a stable partial as an unchanging prefix
that prepends the growing hypothesis. They
modeled the stability statistics of partial
hypotheses using logistic regression or a single-hidden-layer feedforward
network and suggested
emitting only partials that are classified as stable.
In constrast, Selfridge et al.~\cite{selfridge2011stability} defined
partial prefixes as stable only if they prefix the final transcription hypothesis.
They reported the percentage of stable partials by generating
partials only at certain nodes in the lattice. Baumann~\cite{baumann2009assessing}
measured the cumulative differences between subsequent partials up to the hypothesis.
When users see incremental changes in transcription, the cumulative
measurement provides better insight into user experience. In a similar vein, a
recent paper in speech-to-translation~\cite{arivazhagan2019re} used erasure.
Erasure measures the number of tokens to be removed from the suffix of the
previous translation partials to produce the next sequence of partials.
The ratio between aggregated erasure and the final
translation length measures instability of the translation system.

To examine the stability of on-device
streaming E2E ASR models, we propose a novel \textbf{set} of metrics
in Section~\ref{Define} that explicitly measure the speech recognizer stability
users perceive. These metrics are simple and intuitive. They can be captured live
without incurring extra latency on user devices. More importantly, this set
reflects both the frequency and span of revisions in the partials of a speech recognizer.
We then analzye the quality, latency and stability of a RNN-T based E2E streaming
speech recognizer.
We look at the impact of training techniques we used for E2E models on
the stability of RNN-T recognizers: mixed-case data, multi-domain training data,
and text normalization training. We categorize the causes of instability in Section~\ref{RNNT}.
Most existing methods to improve ASR stability focus on delaying partials
from the speech recognizer. We follow this trend to analyze the trade-offs between
end-to-end speech recognizer's latencies and stabilty in Section~\ref{Stream}.
In addition, we evaluate our strategies in improving speech recognizer
stability without delaying partial emissions in Section~\ref{TextNorm}.

%% file: Define.tex
We illustrate the way we measure stability with the example in Table~\ref{tab:exampleUPWRUPSR}.
In this contrived example, the user said ``here comma
lived a man who sailed to sea". The final recognizer output is ``Here, lived a man who
sailed to sea". A segment, or partial result, can
contain multiple words. The emitted segments are chronologically indexed. There are 9
segments in total, although the number could change if the frequency of
partial emissions changes.
A segment differs from the previous segment either by growth (addition of words)
or by revision (changes in the previous words). Revisions indicate unstable segments. For
example, segment 2 is a stable segment because it simply grows from segment 1.
Segment 3, however, is an unstable segment because it contains a revision of
the word ``come" into ``comma". Similar to~\cite{arivazhagan2019re,mcgraw2012estimating}, we
consider a partial prefix stable if and only if all future hypotheses contain
the same prefix. In segment 5, for instance, ``a man who" are unstable words
because they follow an unstable word ``Lived". As a result, we count 4
unstable words in segment 6.

\vspace{-3mm}%reduce too much white space before the table
\begin{table}[h]
\centering
\caption{Incremental speech recognition segments ordered by emission
sequences, with unstable words and segment counts to calculate UPWR and UPSR.}
\label{tab:exampleUPWRUPSR}
\begin{tabular}{|l|l|l|l|}\hline
\textbf{Seg}  & \textbf{Streamed Segment Text} & \multicolumn{2}{|c|}{\textbf{\# Unstable}} \\
\textbf{\#} &                                     & \textbf{Word}       & \textbf{Seg} \\\hline
1 &Here                                  &         &       \\\hline
2 &Here come                             &         &       \\\hline
3 &Here comma                            & +1      & +1    \\\hline
4 &Here,                                 & +1      & +1   \\\hline
5 &Here, Lived a man who                 &         &      \\\hline
6 &Here, lived a man who sell            & +4      & +1   \\\hline
7 &Here, lived a man who sell two seeds  &         &      \\\hline
8 &Here, lived a man who sell 2 seeds    & +2      & +1   \\\hline
9 &Here, lived a man who sailed to sea   & +3      & +1   \\\hline
\end{tabular}
\vspace{-2mm}%reduce too much white space
\end{table}
\vspace{-3mm}%reduce too much white space after the table

We measure the instability of partials with unstable partial word ratio (UPWR)
and unstable partial segment ratio (UPSR). UPWR is the ratio of total number of unstable
words in a test corpus to the total number of words in the final hypotheses.
UPSR captures the ratio between the aggregated number of revised segments
and the total number of utterances in a dataset.
The ranges of UPWR and UPSR are $[0, \infty)$. The closer they are to 0, the more
stable the system. UPWR captures the magnitude of the
model's instability in terms of number of words,
while the UPSR measures the frequency of occurences of revisions.

In Table~\ref{tab:exampleUPWRUPSR}, we have a total of 11 unstable words, and 9 words in
the final hypothesis. Therefore UPWR=11/9=1.22. There are 5 unstable segments
over one utterance, so UPSR=5/1=5.0.

%% file: RNNT.tex
\subsection{On-Device Streaming RNN-T}\label{on-device}
We train a streaming RNN-T with either 128 grapheme targets or 4096 word-pieces
targets~\cite{he2019streaming}. The RNN-T model is trained to handle multiple speech recognition
domains~\cite{Sainath2020stream}. In addition, this RNN-T is trained to run
on modern edge devices~\cite{yuan2020}. Since on-device speech recognizers face
serious constraints in latency, memory, and CPU resources, we avoid adding text normalizers
to the speech pipeline. Instead, we include capitalizations, spoken punctuations
and Arabic numerals in the training data, so that the trained RNN-T model
learns to output transcripts with correct text normalization formats.

\subsection{Types of On-Device RNN-T Instabilities}\label{Types}
\input{TypesOf}

\subsection{Factors Impacting Stability in RNN-T}\label{subsection:factor}
We identify factors that adversely impact the stability of
a mixed-case streaming RNN-T model in the following subsections.

\subsubsection{Wordpieces and Spoken Punctuation Tokens}
Word-piece targets in E2E speech models allow models to capture longer
linguistic and phoneme contexts, resulting in better model performance~\cite{rao2017exploring}.
We train a word-piece model (WPM) with n-gram counts obtained from text data to
segment words into sub-word units. WPMs are depicted by Schuster in~\cite{schuster2012japanese}.
These WPM targets do not contain spoken punctuation tokens.
For example, ``exclamation" is represented by sub-word units as \{``exclam",
``a", ``tion"\}. As a result, the RNN-T models emit partial
punctuation words like ``exclam" before completely converting the spoken
punctuations ``exclamation mark" into ``!".

\subsubsection{Mixed-case Training Data and WPM Targets}\label{subsubsection:mixcase}
In Section~\ref{on-device}, we explain why we introduce mixed-case training data to RNN-T models.
In terms of WER, RNN-T with WPM targets have lower WER than RNN-T with grapheme
targets. In terms of stability, mixed-case training inevitably increases the perplexity
of the RNN-T decoder. RNN-T models with WPM targets suffer greater stability degradations
than RNN-T models with grapheme targets when the training data changes from lowercase to mixed-case.
Table~\ref{tab:wpmvsgrapheme} shows stabilities of Model A (grapheme) and B (WPM)
with lowercase traing data. They have similar stability.
However, RNN-T Model D, trained with a mixed-case WPM,
is 31.3\% worse in stability than its grapheme-trained counterpart C.
Comparing grapheme model pairs C and A, the grapheme-trained RNN-T
has $>40\%$ increase in UPWR and UPSR. Comparing the WPM models B and D,
the RNN-T experiences a 61.5\% increase in UPWR and 90.7\% increase in UPSR.

One might surmise that mixed-case data mainly contributes to an increase in
capitalization instability. Evidence suggests otherwise: mixed-case training causes
other types of instability in WPM-target RNN-T.
To show this, we post-process the outputs of models C and D with a
lowercasing Finite State Transducer (FST). This FST converts all partial
outputs into lowercase text, eliminating capitalization instability completely.
The newly augmented models are ~\textit{Cnorm} and~\textit{Dnorm}
in Table~\ref{tab:wpmvsgrapheme}. \textit{Cnorm} has the same level of
instability as Model A, suggesting that for grapheme RNN-T models,
the $> 40\%$ increase in mixed-case instability is due to capitalization
instabilities alone. \textit{Dnorm}, however, has $30.2\%$ higher UPSR
than Model B. We find that the mixed-case decoding beams of the WPM-targeted
RNN-T often contain more words with both upper and lower cases
than the grapheme-targeted RNN-T. For example, ``used" and ``Used" can both
exist in two otherwise identical top-N hypotheses because both are valid words.
As a result, the diversity of the top N hypotheses decreases,
and the top-1 hypothesis is more likely to
change during the decoding process when the top-n hypotheses swap their rankings.

\subsubsection{Multi-domain Training Data}\label{subsection:multidomain}
Multi-domain training data improves the quality
of the RNN-T model. It supplies a variety of acoustics and linguistic
content to the speech model, improving model performance~\cite{narayanan2018toward}.
The cost of multi-domain training data is that the transcription standard may
not be the same across all domains. Capitalizations, for
instance, might not be enforced in the YouTube transcriptions.
As the diversity of transcription format increases, multi-domain training leads
to more model instability. In Table~\ref{tab:result}, we compare two models:
D, trained with data from a single domain,
and E, trained with data from multiple domains. Although multi-domain
RNN-T model E has a better word error rate (WER) than model D,
its punctuation error rate (PER) increases by 10.7\%,
case-insensitive WER increases by 1.5\%, and its instability metrics increase even more
drastically: UPWR $+76.2\%$ and UPSR $+45\%$.

\begin{table}[hb]
\caption{Stability of WPM and grapheme RNN-T models with training data from a
single domain. ``Grph" is abbreviation for grapheme.}
\label{tab:wpmvsgrapheme}
\begin{tabular}{|l|ll|l|l|l|}
\hline
\textbf{Model}                & \multicolumn{1}{l|}{\textbf{Vocab}} & \textbf{Training} & \textbf{UPWR}        & \textbf{UPSR}         \\
\textbf{Index}                & \multicolumn{1}{l|}{}               & \textbf{Data}     &                        &                       \\ \hline
A                             & \multicolumn{1}{l|}{Grph}           & lowercase         & 0.11                 & 0.48                  \\ \hline
B                             & \multicolumn{1}{l|}{WPM}            & lowercase         & 0.13                 & 0.43                  \\ \hline
C                             & \multicolumn{1}{l|}{Grph}           & mixed-case        & 0.16                 & 0.68               \\\hline
D                             & \multicolumn{1}{l|}{WPM}            & mixed-case        & 0.21                 & 0.82                  \\\hline
\textit{Cnorm}                    & \multicolumn{2}{l|}{Model C}                        & 0.11                 & 0.47                  \\ % this is Gboard 0 + FST
                 & \multicolumn{2}{l|}{+ lowercaseFST}              & 0\% wrt A         & -2.08\% wrt A         \\ \hline
\textit{Dnorm}                    & \multicolumn{2}{l|}{Model D}    & 0.13              & 0.56                  \\ % this is soda v1 model + FST
                & \multicolumn{2}{l|}{+ lowercaseFST}               & 0\% wrt B         & +30.2\% wrt B         \\ \hline
\end{tabular}
\vspace{-2mm}%reduce too much white space
\end{table}

%% file: TypesOf.tex
We analyze the streaming RNN-T output and divide the occurrences of
instability into two categories.
The first is text-normalization instability. The second is streaming
instability. We further identify 4 subtypes of text-normalization
instability. Table~\ref{tab:types} shows the percentage breakdown of each instability subtype in terms
of the frequency of occurrences in the output of the multi-domain RNN-T model,
introduced in~\cite{narayanan2018toward}, which corresponds to Model F in Table~\ref{tab:result}.

\vspace{-2mm}%reduce too much white space
\begin{table}[h]
\centering
  \caption{Types of instability and their percentages of occurrences in an offline dataset.}
  \label{tab:types}
  \begin{tabular}{|l|l|}\hline
    \textbf{Type of Stability}    & \textbf{Percentage}   \\\hline
\textbf{A. Text Normalization Instabilities} &  \textbf{47.6\%} \\\hline
 1,2. Punctuation \& spacing related    & 21.2\%    \\\hline
 3. Capitalization        & 24.7\%                \\\hline
 4. Numeral             & 1.7\%                 \\\hline
 \textbf{B. Streaming instability}      & \textbf{52.4}\% \\\hline
  \end{tabular}
  \vspace{-2mm}%reduce too much white space
\end{table}
\vspace{-4mm}%reduce too much white space

\textbf{Streaming instability} When the model is forced to output partial
words faster, more premature partials occur when the user is still in the process of
uttering a word. For example, one might see ``my open" before ``my opinion".
We analyze the relationship between streaming
instability and partial emissions rate in section~\ref{Stream}. The streaming
instability problem is more pronounced when the E2E models output subword
targets instead of word targets as in the traditional models.

\textbf{Text normalization instabilities} arise when the RNN-T model revises its
transcripts in terms of the formats of the output.
\begin{enumerate}[wide, labelindent=10pt]
\itemsep0em
\item \textbf{Punctuation instability} refers to the change of partials related to spoken
punctuation phrases. Examples include segment 2 to 4 in Table~\ref{tab:exampleUPWRUPSR},
where ``come"$\rightarrow$``comma"
$\rightarrow$``," are partial result revisions due to the comma symbol.
Punctuation instability gets worse as the punctuation phrases get longer.
``left quotation mark", for example, causes more instability than ``period".
\item \textbf{Spacing instability} refers to the changes in spaces delimiting the partial words during the
recognition process. It often happens hand-in-hand with punctuation instability.
A commonly occurring observation is that the space between a word and the
subsequent punctuation is being removed and re-inserted multiple times:
``Hi," $\rightarrow$ ``Hi ," $\rightarrow$ ``Hi, ". Languages that do not
require delimitation, such as Chinese and Japanese, have no spacing instabilities.
\item \textbf{Capitalization instability} is caused by the model revising uppercase
outputs into lowercase outputs or
vise-versa. Segments 5 to 6 in Table~\ref{tab:exampleUPWRUPSR} are
examples of capitalization instability (i.e. ``Lived" $\rightarrow$ ``lived").
Capitalization instability is more pronounced in languages like German,
where nouns are capitalized.
\item \textbf{Numeral instability} occurs when users dictate phone numbers,
street addresses, dates, time or other numeric entities. When the user is
speaking, the ASR models output these
numbers in spoken format first - ``Call eight" - before outputting numbers
- ``Call 800-123-1234".
\end{enumerate}

%% file: TextNorm.tex
In this section, we discuss methods that improve E2E RNN-T recongizer stability
without introducing a delay in partial emissions. We develop strategies
to reap the benefits of improved WER while alleviating the degradations of
model stability explained in section~\ref{subsection:factor}.

\begin{table*}[th]
  \caption{RNN-T model quality and stability. PEI: partial emission interval (ms);
  PER: punctuation error rate; mWER: mixed-case WER.}
  \label{tab:result}
  \centering
  \begin{tabular}{|l|l|l|l|l|l|l|l|l|}\hline
  \textbf{ID} &  \textbf{Vocab} &\textbf{PEI}& \textbf{Training Domain \& Data}  & \textbf{WER}  &\textbf{PER}& \textbf{mWER}         & \textbf{UPWR} & \textbf{UPSR} \\\hline
  C           &       Grapheme  &50& Single                            & 4.7(0\%)                &    2.5     & 6.9(0\%)                & 0.16          & 0.68       \\\hline %V0 Gboard baseline 4.7 WER, 6.9 mWER
  D           &       WPM       &50& Single                            & 4.3(-8.51\%)            &    2.8     & 6.6(-4.3\%)             & 0.21          & 0.82          \\\hline  %V1 SODA 4.3 WER, 6.6mWER
  E           &       WPM       &50& Multi                             & 4.0(-14.9\%)            &    3.1     & 6.7(-2.9\%)             & 0.37          & 1.19       \\\hline %V2 or V3 Soda 4.0 WER, 6.7mWER
  F           &       WPM       &50& Multi + Text Norm                 & 3.8(-19.1\%)            &    2.6     & 6.2(-10.1\%)           & 0.26          & 0.96       \\\hline % V9 Gboard 4.0 WER (PER relies on the re-trained model), 6.3 mWER
  G           &       WPM       &50& Multi + Text Norm + Domain-id     & \textbf{3.6(-23.4\%)}   &\textbf{2.3}&\textbf{5.8(-15.9\%)}  & 0.14          & 0.48       \\\hline %22 Gboard 3.6 WER, 5.8mWER
  H           &       WPM       &200& Multi + Text Norm + Domain-id    & \textbf{3.6(-23.4\%)}   &\textbf{2.3}&\textbf{5.8(-15.9\%)}  & \textbf{0.04} & \textbf{0.15}   \\\hline %22 Gboard 3.6 WER (400ms partial emission)
  \end{tabular}
\vspace{-3mm}%reduce too much white space
\end{table*}

\subsection{Numeral Instability}\label{numerics}
To reduce numeral instability, we included
Text-to-speech (TTS) synthesized number data as introduced in~\cite{he2019streaming}
to fine-tune the models so that the RNN-T models automatically output the
correct numeral formats.

\subsection{Punctuation Words as WPM Tokens}\label{wpmtokenpunct}
To eliminate punctuation instabilities, we force the RNN-T models to predict punctuation
phrases as single tokens. We add a list of possible punctuation
phrases in the single-token format, such as ``\{exclamation-mark\}" or ``\{left-curly-bracket\}".
We then add these tokens in the WPM vocabulary.
At training time, we pre-process
the audio transcriptions to ensure that dictated punctuations occur
in the same token format. The trained ASR model learns to distinguish common
words from punctuation phrases in their linguistic context; it predicts ``period" when it means
a stretch of time and ``\{period\}" when it means end-of-sentence. At inference
time, punctuation words with curly brackets are instantly converted into symbols
by a simple regular expression logic.

\subsection{Text Normalization and Domain-id}\label{subsection:solution}
To alleviate the capitalization instability in mixed-cased trained RNN-T
models, we implement two solutions: text normalization on the multi-domain dataset,
and separation of domains using domain-id as a model input feature.

We apply text normalization models to the multi-domain training
data~\cite{chua2018text} to unify the format of capitalization, spoken
punctuations, numerics, and spacing clean-ups. Comparing Model E to F in
Table~\ref{tab:result}, text normalization improves the
stability of multi-domain model by $29.7\%$ in UPWR and $19.3\%$ in UPSR.

We use domain-id as input to the RNN-T model as explained in~\cite{Sainath2020stream}.
Domain-id allows the model to distinguish different text normalization standards
and learn a more consistent text format from each domain.
Model G with domain-id shows about $50\%$ stability improvement over Model F.

With text normalization and domain-id, we develop Model G, a mixed-case streaming
RNN-T with WPM targets. It has $23.4\%$ better WER, $15.9\%$ better
case-insensitive WER, $8\%$ better punctuation error rate and $12.5\%$ improved UPWR
than our baseline mixed-case single-domain grapheme-based RNN-T (Model C).

%% file: Stream.tex
\subsection{Stability vs Partial Emission Intervals}
Previous works have shown trade-offs between partial emission intervals (PEI) and model
stability~\cite{fink1998incremental,baumann2009assessing,mcgraw2012estimating,arivazhagan2019re}.
PEI is the time we set between showing consequtive partial results from the ASR recognizer.
Naturally, the longer a partial lives in the lattice beam,
the less likely the partial word is going to be revised.

\vspace{-4mm}%reduce too much white space
\begin{figure}[h]
  \centering
  \includegraphics[width=0.95\linewidth]{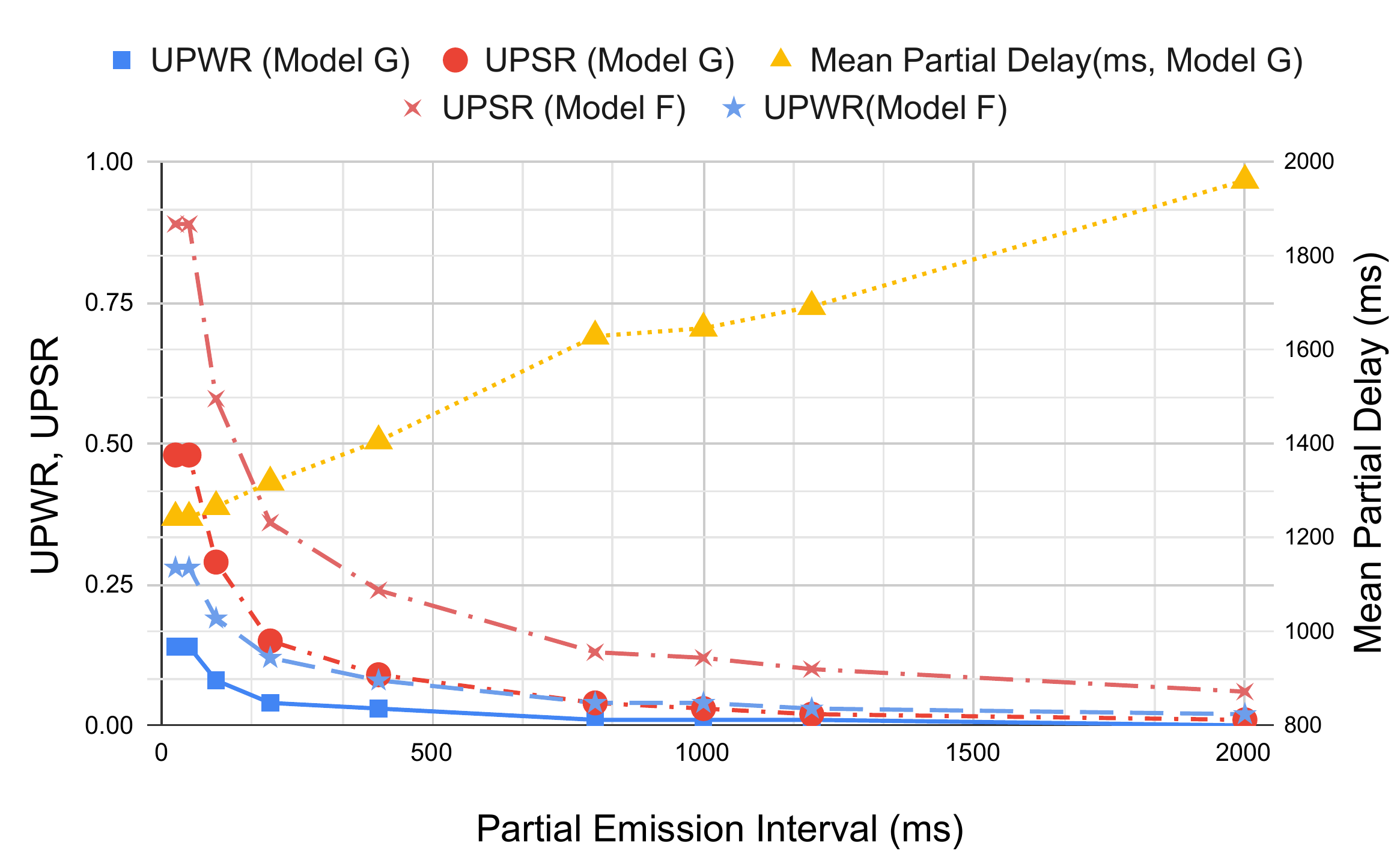}
  \vspace{-2mm}%reduce too much white space
  \caption{Instability (UPWR, UPSR) with respect to partial emission
  intervals (ms) in Model G and Model F. Mean partial delay (ms) of Model G
  is also shown.}
  \label{fig:InstabilityVsPartial}
\end{figure}
\vspace{-4mm}%reduce too much white space

We thus explore the relationship between partial emission intervals (in ms) and model stability
with a streaming RNN-T E2E speech recognizer. Figure~\ref{fig:InstabilityVsPartial} shows how the
instability metrics decrease logarithmically when
the partial emission interval lengthens. In Section~\ref{TextNorm}, we introduce methods to
stablize WPM target RNN-T models by pushing the UPWR and UPSR curves to a lower
magnitude. For instance, by using domain-ID, we show improvement of stability from
model F to model G at no delay of the partial emission intervals at every point along the
UPWR and UPSR curves in Figure~\ref{fig:InstabilityVsPartial}. In this section, we
show how stability can be improved by sliding to different PEI's along the curves.
We also measure the latency of the speech recognizer in terms of mean partial delay,
which is the average time before each hypothesized word is first shown on the screen after the
beginning of the utterance.

Changing the partial emission interval from 50ms to 200ms results in 75ms increase in mean partial
delay (see Figure~\ref{fig:InstabilityVsPartial}). However, at 200ms PEI, as shown in
Table~\ref{tab:result}, model H has $71.6\%$ improvement in UPWR, and $68.8\%$ improvement in UPSR,
compared to model G, achieving a good tradeoff between latency and stability.

\subsection{Stability Thresholding vs Mean Partial Delay Tradeoff}
\begin{figure}[h]
  \centering
  \includegraphics[width=0.95\linewidth]{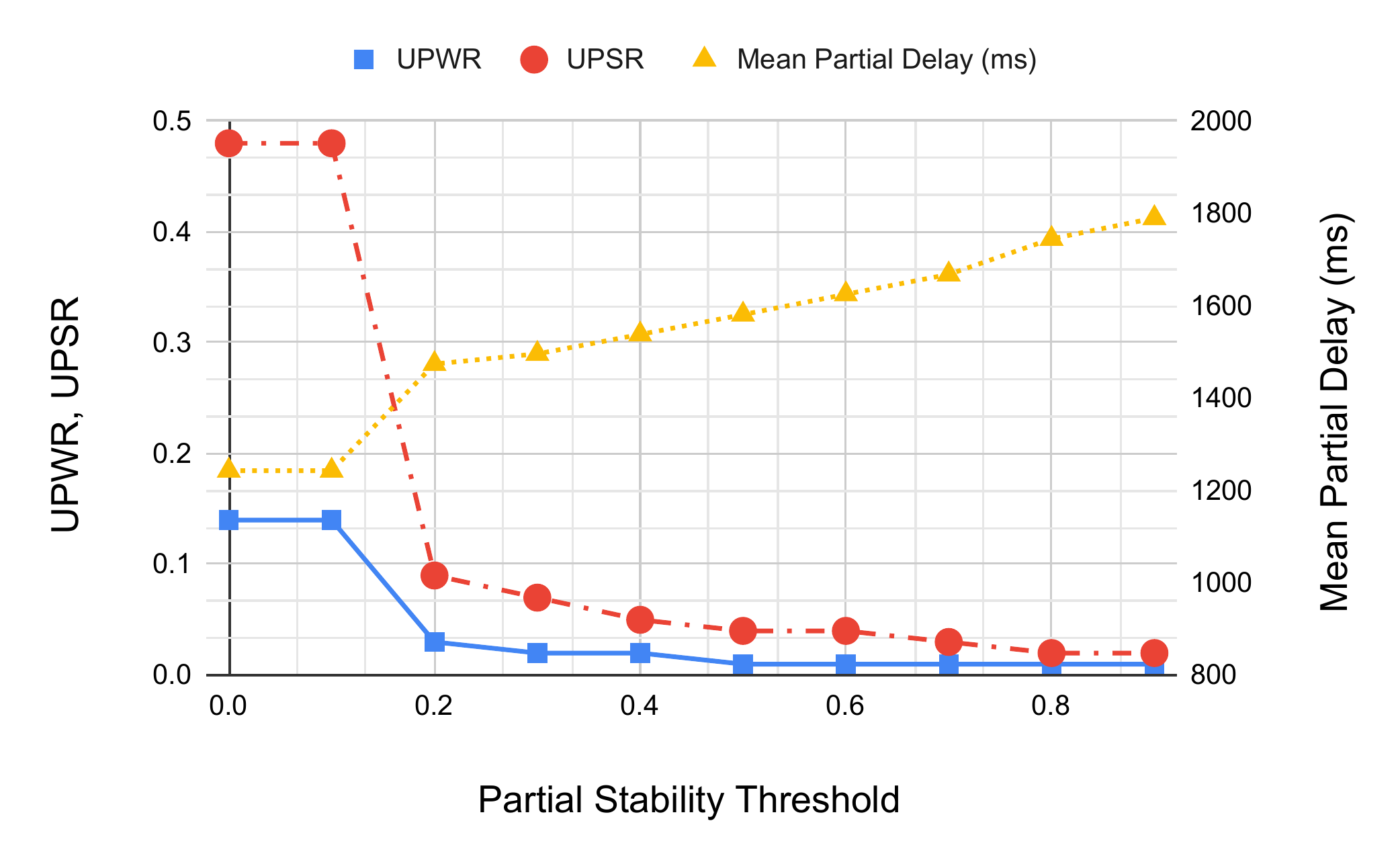}
  \vspace{-2mm}%reduce too much white space
  \caption{UPWR, UPSR and mean partial delay with respect
  to the threshold of stability score in Model G.}
  \label{fig:InstabilityVsThreshold}
\end{figure}

In~\cite{mcgraw2012estimating}, McGraw and Gruenstein developed a logistic
regression approach to estimate the stability of partial results. We analyze the
impact on model stability by thresholding the stability scores of partials.
Each partial is scored based on
the proposed logistic regression model, and partial
words that have stability score exceeding the threshold are immediately shown on the
screen while the others are withheld. Figure~\ref{fig:InstabilityVsThreshold}
shows that increasing the threshold improves
model stability. UPWR and UPSR drop sharply when the threshold grows
from 0.1 to 0.2, and decrease linearly when the threshold score is bigger than 0.2.

Accompanying the precipitous drop in the speech recognizer's UPSR and UPWR is a
perceptible increase in recognizer latency. This is measured by a $18.6\%$
increase of the mean partial delay from 1243ms to 1474ms. This delay is expected
because the logistic regression model depends on a feature, \textbf{age}, which
measures the length of time that a partial survives the best decoding path.
Partials at the beginning of an utterance usually do not have long \textbf{age},
and thus are predicted to be less stable.

\vspace{-4mm}%reduce too much white space
  \vspace{-6mm}%reduce too much white space

%% file: Conclude.tex
In this paper, we describe the problem of instability
in the framework of a streaming E2E RNN-T based multi-domain ASR.
We observe that model training methods such as mixing data from multiple
speech domains and adopting WPM-targets have adversely impacted on the
stability of the streaming RNN-T models.
In addition, the introduction of text normalization
features, including mixed-case data, numerics and spoken punctuations, has
also resulted in model stability degradation.
We first introduce a novel set of metrics, UPWR and UPSR, to quantify
the magnitude and frequency of instability occurences.
We then categorize instability into 5 types. We
outline model training techniques that dramatically improve the stability of
streaming E2E models at no delay to the output words.
We show partial emission delay is an effective tool to reduce streaming
instability of the RNN-T models but only up to a point before the partial
emissions delays become perceptible.